\documentclass[letterpaper, 10 pt, conference]{ieeeconf}  

\IEEEoverridecommandlockouts 
\usepackage{graphicx} \usepackage{times} \usepackage{amsmath}
\usepackage{amssymb}
\usepackage{bm}
\usepackage{mathtools} 
\usepackage{algorithm}
\usepackage{booktabs}
\usepackage[noend]{algpseudocode}
\usepackage{epsfig} \usepackage{epstopdf}
\usepackage{multirow} \usepackage{subfigure}
\usepackage{caption} \usepackage{cite}
\usepackage{color}	\usepackage{url} 
\usepackage{gensymb} 
\usepackage{balance}

	\newtheorem{example}{Example}

	\DeclareMathOperator*{\argmin}{arg\,min}

\renewcommand{\baselinestretch}{0.97}

\title{
Self-Supervised Motion Retargeting with Safety Guarantee 
}
\author{Sungjoon Choi$^{\dag}$, 
Min Jae Song$^{\ddag}$,
Hyemin Ahn$^{\S}$,
and Joohyung Kim$^{\P}$
\thanks{$^{\dag}$Sungjoon Choi is with 
School of Artificial Intelligence, 
Korea University, Seoul, Korea
{\tt\footnotesize sungjoon-choi@korea.ac.kr}.}%
\thanks{$^{\ddag} $Min Jae Song is with 
Courant Institute of Mathematical Sciences, 
New York University, New York, NY, USA 
{\tt\footnotesize minjae.song@nyu.edu}.}%
\thanks{$^{\S} $Hyemin Ahn is with
Chair of Human-centered Assistive Robotics, 
Technical University of Munich, Munich, Germany
{\tt\footnotesize hyemin.ahn@tum.de}.}%
\thanks{$^{\P} $Joohyung Kim is with 
University of Illinois at Urbana-Champaign, 
Champaign, IL, USA
{\tt\footnotesize joohyung@illinois.edu}.}%
}
\begin{document}
\maketitle 

%
%
\begin{abstract}
In this paper, we present self-supervised shared latent embedding (S$^3$LE),
a data-driven motion retargeting method
that enables the generation of natural motions in humanoid robots from 
motion capture data or RGB videos.
While it requires paired data consisting of human poses and
their corresponding robot configurations,
it significantly alleviates the necessity of time-consuming 
data-collection via novel paired data generating processes.
Our self-supervised learning procedure consists of two steps:
automatically generating paired data to bootstrap the motion retargeting,
and learning a projection-invariant mapping to handle 
the different expressivity of humans and humanoid robots.
Furthermore, our method guarantees that the generated robot pose is 
collision-free and satisfies position limits
by utilizing nonparametric regression in the shared latent space.
We demonstrate that our method can generate 
expressive robotic motions from both the CMU motion capture database
and YouTube videos.
\end{abstract}
\IEEEpeerreviewmaketitle 

%
%
\section{Introduction}

Generating natural and expressive robotic motions for humanoid robots 
has gained considerable interest from both robotics and computer graphics
communities~\cite{Gleicher_98,Lawrence_04,Shon_06,Levine_12_Character,Huang_18}.
Recently, this phenomenon has been accelerated by the fact that 
more human-like robots are permeating our daily lives
through applications such as interactive services or educational robots.
However, in order to generate a number of natural motions for humanoid robots, 
a substantial amount of effort is often required to manually design 
time-stamped robotic motions by animators or artists.

One alternative approach is to leverage existing motion capture
or animation data to generate robotic motions, which is often referred to 
as motion retargeting. 
Traditionally, motion retargeting is performed by manually defining 
a mapping between two different morphologies 
(e.g., a human actor and an animation character).
This requires one to 
first design (or optimize) the pose feature of the source domain
to transfer, and then find the corresponding pose in the target domain.
For instance, the source and target can be human skeletons and 
humanoid robot joint angles, respectively.
However, a substantial amount of effort is often required to 
manually design a proper mapping between the domains 
because kinematic constraints have to be taken into consideration as well.

On the other hand, data-driven motion retargeting has been used
to circumvent the manual mapping process by leveraging
machine learning methods~\cite{Lawrence_04}. 
Such learning-based methods enjoy flexibility and scalability
as they reduce the need for excessive domain knowledge
and tedious tuning processes required to define pose features properly.
Another important benefit of data-driven motion retargeting is 
the lightweight computation requirement of the execution phase
as no iterative optimization is involved during the execution. 
However, one clear drawback is that we have to collect 
a sufficient number of training data in advance.
Moreover, it is not straightforward how to ensure the feasibility 
of the motion when using a learning-based model.

In this paper, we present self-supervised shared latent embedding (S$^3$LE),
a data-driven motion retargeting method
that enables the generation of natural motions in humanoid robots from 
motion capture data or RGB videos.
In particular, we apply self-supervised learning in two domains:
1) generating paired data for motion retargeting without relying on the 
human motion data 
and 
2) learning a projection-invariant mapping 
to handle the different expressivity of humans and humanoid robots. 
The proposed data generating process not only alleviates the necessity of
collecting a large number of pairs consisting of 
human pose features and corresponding robot configurations, but 
also allows the motion retargeting algorithm not to be restricted
to the given motion capture dataset.
Furthermore, the proposed projection-invariant mapping, 
which is defined as a mapping from an arbitrary human skeleton 
to a motion-retargetable skeleton, 
takes into account the fact that the range of robot poses 
are often more restrictive than those of humans'.
By combining the recently proposed motion retargeting method~\cite{Choi_2020_RSS}
that can guarantee the feasibility of the transferred motion, 
our proposed method can perform motion retargeting online 
with safety guarantee 
(i.e., self-collision free) under mild assumptions. 



%
%
\section{Related Work}

In this section, we give a summary of existing work 
on data-driven motion retargeting 
and self-supervised learning methods in robotics, 
which form the foundation of this work. 

%
%
\subsection{Data-driven Motion Retargeting}
Data-driven motion retargeting methods have been widely used 
due to its flexibility and scalability
\cite{Lawrence_04,Shon_06,Levine_12_Character,Huang_18}. 
Owing to the merits of such methods, \cite{Yamane_10} 
was able to generate convincing motions of non-humanoid characters 
(i.e., a lamp and a penguin) from human MoCap data. 
Many data-driven techniques have relied on
Gaussian process latent variable models (GPLVM) 
to construct shared latent spaces 
between the two motion domains \cite{Lawrence_04, Shon_06}. 
More recently, \cite{Yin_17} proposed associate latent encoding (ALE),
which uses two different variational auto-encoders (VAEs) 
with a single shared latent space.
A similar idea was extended in \cite{Choi_20_RAL}
by incorporating negative examples to aid safety constraints. 
However, these methods do not consider the different expressivity
of humans and robots. That is, multiple human poses can be
mapped into a single robot configuration due to physical constraints.
In this regard, we present the projection-invariant mapping 
while constructing the shared latent space. 

%
%
\subsection{Self-supervised Learning in Robotics}
Self-supervised learning has found diverse applications in robotics 
such as learning invariant representations of visual inputs~\cite{Sermanet_17_tcn,Dwibedi_18_learning,Laskin_20_curl}, 
object pose estimation~\cite{Deng_20_self}, 
and motion retargeting~\cite{Villegas_18,Lim_19_pmnet}.
The necessity of self-supervision arises naturally in robotics, 
where every piece of data often requires the execution of a real system.
Collecting large-scale data for robotics poses significant challenges
as robot execution can be slow and producing high-quality labels can be 
expensive. 
Self-supervision allows one to sidestep this issue 
by generating cheap labels (but with weak signals) for vast amounts of data. 
For instance,~\cite{Gandhi_17_learning} creates a large dataset of 
Unmanned Aerial Vehicle (UAV) crashes, and labels an input image as positive 
if the UAV is far away from a collision and negative otherwise. 

Moreover, self-supervised learning can be used to train representations 
that are invariant to insignificant changes, 
such as change of viewpoint and lighting in image inputs~\cite{Sermanet_17_tcn}. 
As for self-supervised learning methods for motion retargeting, 
which is the problem of our interest,~\cite{Villegas_18} and~\cite{Lim_19_pmnet} 
train motion retargeting networks with \emph{unpaired} data using 
a cycle-consistency loss~\cite{Zhu_17cyclegan} in either the input or the latent space. 
While both methods do impose a soft penalty for violating 
joint limits during the training procedure, 
they do not guarantee feasibility as the joint limits 
are not strictly enforced, and self-collision is not taken into consideration.

%
%
\section{Preliminary}
Considering the safety and feasibility of 
the retargeted robotic motion is crucial in robotics. 
In particular, when we are using a learning-based model
using a function approximator (e.g., neural networks), 
it is not straightforward to guarantee that the 
resulting prediction lies within the safety region. 
In this regard, the recently proposed 
locally weighted latent learning (LWL$^2$)~\cite{Choi_2020_RSS}
performs nonparametric regression 
on a shared deep latent space. 
One unique characteristic is that one can always
guarantee the feasibility of the mapping by finding the
closest feasible point in the latent space. 
Our method also enjoys this property by leveraging LWL$^2$. 

Suppose we want to find a mapping from
$\mathbb{X}$ to $\mathbb{Y}$ using 
$N$ paired data with correspondences
(i.e., $\mathcal{D} = \{ \mathbf{x}_i, \mathbf{y}_i \}_{i=1}^N$).
The core idea of our method is to construct a shared latent space
$\mathbb{Z}$ by learning appropriate encoders and decoders of 
$\mathbb{X}$ and $\mathbb{Y}$ (i.e.,
$Q^{\mathbb{X}}: \mathbb{X} \to \mathbb{Z}$,
$Q^{\mathbb{Y}}: \mathbb{Y} \to \mathbb{Z}$,
$P^{\mathbb{X}}: \mathbb{Z} \to \mathbb{X}$, and
$P^{\mathbb{Y}}: \mathbb{Z} \to \mathbb{Y}$).
Both encoder networks $Q$ and decoder networks $P$ are trained 
using the Wasserstein autoencoder (WAE) \cite{Tolstikhin_18} framework
with the following loss function in domain $\mathbb{X}$ (and $\mathbb{Y}$):
\begin{align} \label{eqn:wae_loss}
\begin{split}
  \mathcal{L}_{\text{WAE}}
    &=
    \frac{1}{N}
      \sum_{i=1}^N
        c(\mathbf{x}_i,P^{\mathbb{X}}_{\theta}
          (Q^{\mathbb{X}}_{\phi}(\mathbf{x}_i)))
    \\
    &\quad - 
    \frac{\beta}{2}
    \left[
      \log( \rho(D_{\psi}(\mathbf{z}_i)) )
      + \log
        \left( 
          1 - \rho(D_{\psi}(Q^{\mathbb{X}}_{\phi}(\mathbf{x}_i))) 
        \right)
    \right]|_{\psi} 
    \\
    &\quad -
    \beta
    \left[
      \log(\rho(D_{\psi}(Q^{\mathbb{X}}_{\phi}(\mathbf{x}_i))))
    \right]|_{\phi}
\end{split}     
\end{align}
where
$N$ is the number data, 
$D_\psi: \mathbb{Z} \to \mathbb{R}$ is a discriminator for latent prior fitting,
$c(\cdot)$ is a distance function defined in the input space,
$\beta$ controls the latent prior fitting (we default to $\beta=1$),
$\rho(\cdot)$ is the sigmoid function, and
$|_{\phi}$ and $|_{\psi}$ indicate the trainable variables, e.g., 
$|_{\phi}$ indicates that only $\phi$ is being updated\footnote{
Please refer \cite{Tolstikhin_18} for more details.}.

While (\ref{eqn:wae_loss}) can be used to construct a single latent space,
we need an additional loss function to \emph{glue} the two latent spaces.
To this end, we use the following latent consensus loss:
\begin{equation} \label{eqn:lc_loss}
  \mathcal{L}_\text{lc} = 
    \frac{1}{N_{\text{pair}}} \sum_{i=1}^{N_{\text{pair}}}
    \| Q_{\phi}^{\mathbb{X}}(\mathbf{x}_i)
     -Q_{\phi}^{\mathbb{Y}}(\mathbf{y}_i)
    \|_2^2
\end{equation}
where $N_{\text{pair}}$ is the number of paired data
(i.e., 
$\mathcal{D}_{\mathbb{X}\mathbb{Y}} = 
\{
\mathbf{x}_i, \mathbf{y}_i
\}_{i=1}^{N_{\text{pair}}}$).
Note that one can use domain-specific data of each domain
(without correspondence) for optimizing (\ref{eqn:wae_loss})
and the glue data of both domains ($\mathcal{D}_{\mathbb{X}\mathbb{Y}}$) 
for optimizing (\ref{eqn:lc_loss}).

Once shared latent space is constructed, one can use 
nearest neighbor search in the latent space to find the mapping
between $\mathbb{X}$ and $\mathbb{Y}$.
Given a test input $\mathbf{x}_* \in \mathbb{X}$,
the corresponding test output $\mathbf{y}_*$ can be computed
in a nonparametric fashion as
\begin{equation} \label{eqn:dlu} 
  \mathbf{y}_* = \argmin_{\mathbf{y} \in \mathcal{D}_{\mathbb{Y}}}
    d_{\mathbb{Y}} \left( 
      Q^{\mathbb{X}}(\mathbf{x}_*), Q^{\mathbb{Y}}(\mathbf{y}) 
      \right)
\end{equation}
where $\mathcal{D}_{\mathbb{Y}}$ is a set of points in 
$\mathbb{Y}$.
We would like to stress that the 
domain-specific data
$\mathcal{D}_{\mathbb{Y}}$ does not necessarily have to be 
identical to the ones in $\mathcal{D}_{\mathbb{X}\mathbb{Y}}$
but rather can be easily augmented from sampling points in
$\mathbb{Y}$ as we do not have to care about correspondence issues
with points in $\mathbb{X}$.

In our motion retargeting problem, 
$\mathbb{X}$ and $\mathbb{Y}$ correspond to the space
of human skeletons and robot joint configurations, respectively.
Similarly, $\mathcal{D}_{\mathbb{X}\mathbb{Y}}$ and $\mathcal{D}_{\mathbb{Y}}$ 
correspond to the pairs of human skeletons and 
motion retargeted robot configurations,
and feasible robot configurations (not paired with human skeletons), 
respectively. 
Thus, collecting $\mathcal{D}_{\mathbb{Y}}$
is much easier than collecting $\mathcal{D}_{\mathbb{X}\mathbb{Y}}$.

%
%
\begin{figure*}[t!]
  \centering
  \includegraphics[width=0.95\textwidth]{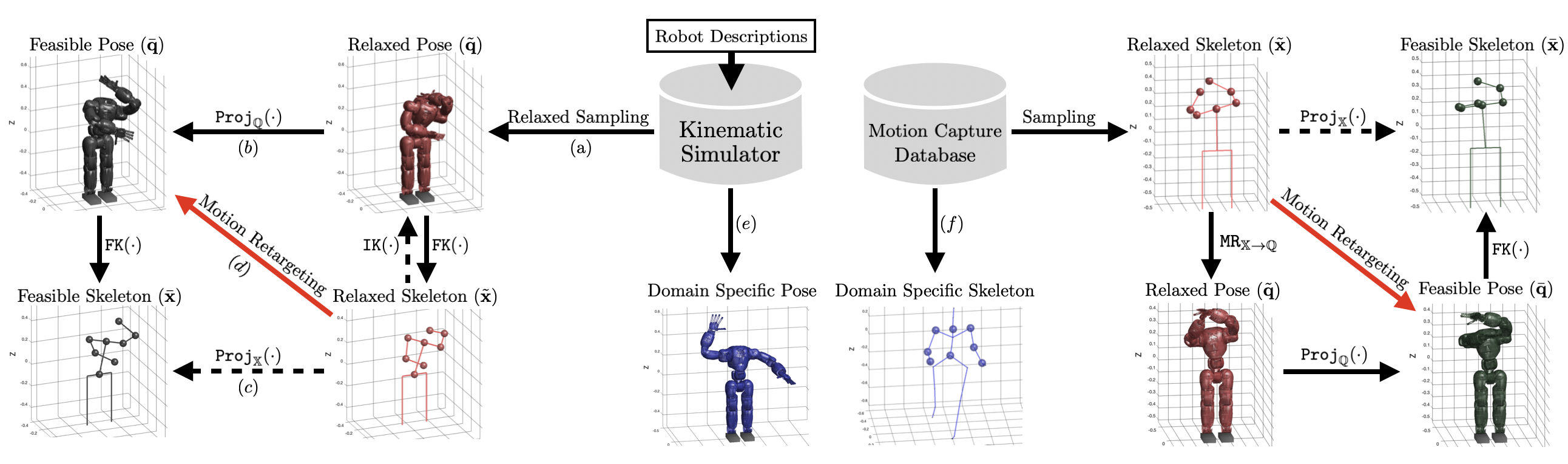}
  \vspace{-0.1pt}
  \caption{
  An overview of the proposed sampling pipelines.
  From the kinematic information of a robot, 
  (a) a relaxed pose $\widetilde{\mathbf{q}}$ is sampled by relaxing 
  the kinematic constraints such self-collision violations. 
  A feasible pose 
  $\overline{\mathbf{q}} = \texttt{Proj}_{\mathbb{Q}}(\widetilde{\mathbf{q}})$
  of a robot is computed by (b) a pose projection mapping where 
  $\overline{\mathbf{q}}$ always satisfies the feasibility conditions. 
  Relaxed and feasible skeleton features 
  $\widetilde{\mathbf{x}}$ and $\overline{\mathbf{x}}$ are collected
  by first solving forward kinematics of
  $\widetilde{\mathbf{q}}$ and $\overline{\mathbf{q}}$ and 
  computing the skeleton features, respectively, where 
  (c) a skeleton projection mapping 
  $\texttt{Proj}_{\mathbb{X}}: \widetilde{\mathbf{x}} \mapsto \overline{\mathbf{x}}$
  is implicitly defined. 
  Also, (e) robot-specific and (f) mocap-specific data are collected. 
  }
  \label{fig:overview}
  \vspace{-0.1pt}
\end{figure*}

%
%
\section{Self-Supervised Motion Retargeting}

In this section, we present our self-supervised 
motion retargeting method.
The cornerstones of our method are
paired data sampling processes in Section~\ref{subsec:sample_from_mocap}
and Section~\ref{subsec:sample_from_robot}
for generating a sufficient number of tuples with correspondences.
Our main motivation is to incorporate
the different expressivity of humans and humanoid robots.
In other words, due to the joint limits and physical constraints of robots,
the mapping from a human skeleton to a robot pose 
may not be a one-to-one mapping, 
but rather an $n$-to-one mapping since multiple human poses can be mapped
to a single robot configuration.
Self-supervised learning with
normalized temperature-scaled cross-entropy
is utilized to handle this different expressivity. 

\subsection{Problem Formulation} \label{subsec:prob_form}

We first introduce basic notations. 
We denote by $\mathbf{x}$ and $\mathbf{q}$ 
the human skeleton representation
and robot joint configuration, respectively. 
Specifically, 
$\mathbf{x} \in \mathbb{X} \subset \mathbb{R}^{21}$ 
is a $21$-dimensional vector consisting of seven unit vectors:
hip-to-neck, both left and right 
neck-to-shoulder, shoulder-to-elbow, and elbow-to-hand unit vectors.
A robot configuration $\mathbf{q} \in \mathbb{Q} \subset \mathbb{R}^{N_{\text{DoF}}}$ 
is a sequence of $N_{\text{DoF}}$ revolute joint angles\footnote{
This could limit the motion retargeting results in that orientation 
information is omitted. However, this makes the proposed method
compatible with the output of pose estimation algorithms for RGB videos. 
}. Moreover, we denote relaxed skeletons and robot configurations that do not necessarily satisfy the robot's physical constraints by $\tilde{\mathbf{x}}$ and $\tilde{\mathbf{q}}$, and feasible skeletons and configurations by $\bar{\mathbf{x}}$ and $\tilde{\mathbf{q}}$.
Throughout this paper, we use robot poses and joint configurations
interchangeably
as they have one-to-one correspondences.
Our goal of motion retargeting is to find a mapping from 
$\mathbb{X}$ to $\mathbb{Q}$
(i.e., human skeleton space to robot configuration space),

To handle the diverse human poses that map into
a single robot pose, we leverage self-supervised learning
for constructing the shared latent space. 
Specifically, the following normalized temperature-scaled cross entropy 
(NT-Xent)
loss will be used: 
\begin{align*}
  \ell(i,j) = -\log\frac{\exp(\cos(\mathbf{z}_i',\mathbf{z}_j')/\tau)}{\sum_{m \neq i} \exp(\cos(\mathbf{z}_i',\mathbf{z}_m')/\tau)}
\end{align*}
\begin{align*}
  \mathcal{L}_\text{nt-xent} = \frac{1}{2N} \sum_{m=1}^N \ell(2m-1,2m) + \ell(2m, 2m-1)
\end{align*}
where $\mathbf{z}_i'$ is the $i$-th latent vector on latent space
mapped by an encoder network and $\tau$ is a temperature parameter
which we set to be $1$. 
In image classification domains, 
$\mathbf{z}_i'$ and $\mathbf{z}_j'$ correspond to the feature vectors
of two images that are augmented from a single image with different
augmentation methods~\cite{Chen_20_simclr}.
In our setting, $\mathbf{z}_i'$ and $\mathbf{z}_j'$ correspond to latent representations of skeletons that map to the same robot configuration.
We present two sampling processes in 
Sections~\ref{subsec:sample_from_mocap} and~\ref{subsec:sample_from_robot} that
allows us to construct $\mathbf{z}_i'$ and $\mathbf{z}_j'$ pairs
which are encoded from a feasible (retargetable) and 
a relaxed (possibly not retargetable) skeletons, respectively. 
Figure \ref{fig:overview} illustrates the overview of the two sampling processes. 

%
%
\begin{figure}[t!]
  \centering
  \includegraphics[width=0.46\textwidth]{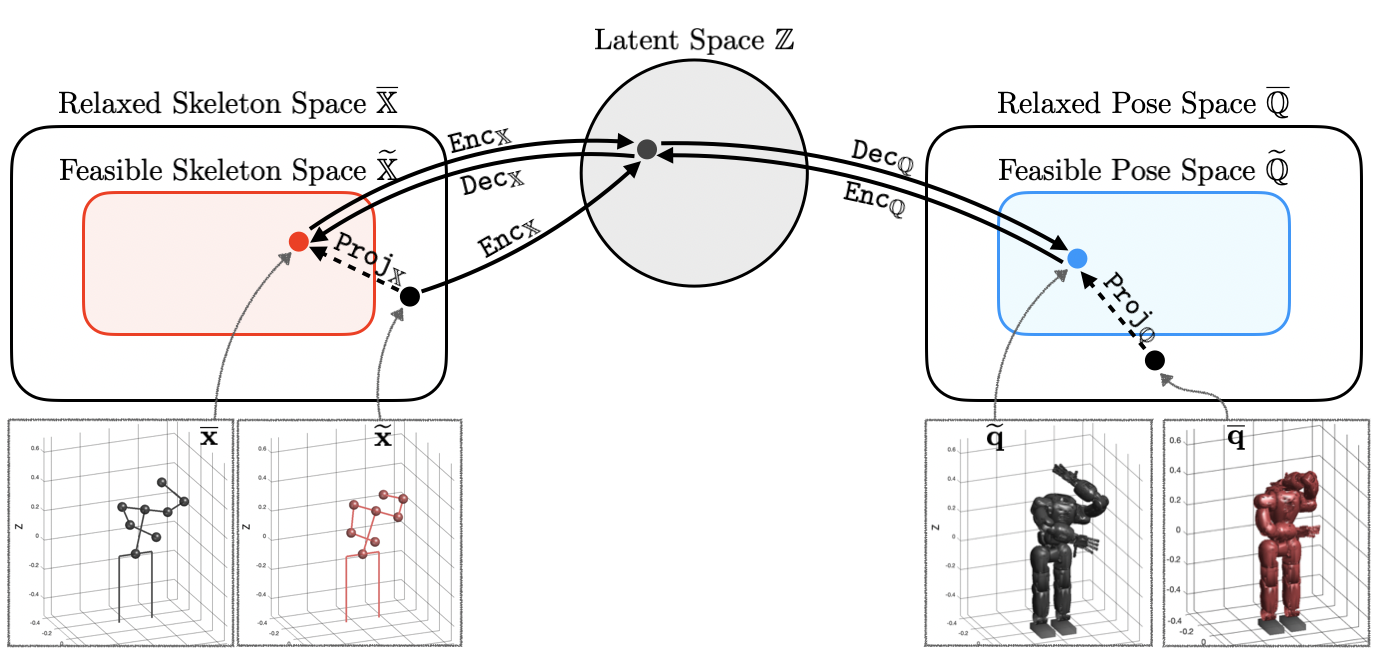}
  \vspace{-0.1pt}
  \caption{
  Projection-invariant mapping 
  from $\mathbb{X}$ to $\mathbb{Z}$.
}
\label{fig:mapping}
\vspace{-5pt}
\end{figure}

%
%
\begin{figure*}[t!]
  \centering
  \includegraphics[width=0.8\textwidth]{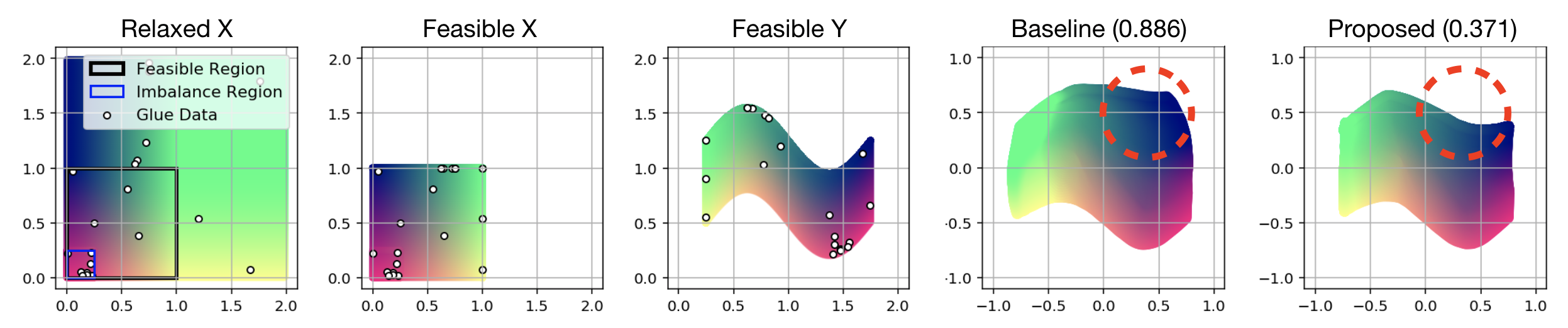}
  \vspace{-0.1pt}
  \caption{
    A synthetic example to validate the benefit of the proposed
    self-supervised learning in constructing 
    the shared latent space.
    Points with the same colors have correspondences with each other. 
  }
  \label{fig:synthetic_transfer}
  \vspace{-0.1pt}
\end{figure*}

%
%
\subsection{Sampling Data from Motion Capture Data} \label{subsec:sample_from_mocap}

Our proposed method still requires some amount of paired data, so 
we use an optimization-based motion retargeting method~\cite{Choi_19}
to create a small paired dataset of human skeletons
and corresponding robot configurations.
However, unlike the approach taken in~\cite{Choi_19}, which simply 
computes pairs of human poses ($\widetilde{\mathbf{x}}$) 
and robot poses ($\overline{\mathbf{x}}$), 
we carefully design an additional sampling routine so that
the resulting tuple can be used for contrastive learning. 

First, we randomly select a human pose from the motion capture dataset
($\widetilde{\mathbf{x}}$) and run 
optimization-based motion retargeting \emph{without} considering 
the joint limits and self-collision, which gives us a relaxed robot pose
($\widetilde{\mathbf{q}}$). 
Note that this process takes considerable
amount of time due to solving inverse kinematics as a sub-routine. 
Then, we compute the feasible robot pose ($\overline{\mathbf{q}}$)
from $\widetilde{\mathbf{q}}$ using the method presented in \cite{Choi_2020_RSS}.
This process can be seen as {\it projecting} 
$\widetilde{\mathbf{q}}$ to $\overline{\mathbf{q}}$
(i.e., $\texttt{Proj}_{\mathbb{Q}}: \widetilde{\mathbf{q}} 
\mapsto \overline{\mathbf{q}}$).
Finally, from $\overline{\mathbf{q}}$,
we compute the skeleton representation by solving forward kinematics
and concatenating appropriate unit vectors described in
Section \ref{subsec:prob_form}
(i.e., $\overline{\mathbf{x}} = \texttt{FK}( \overline{\mathbf{q}} )$).
The collected tuple 
($\widetilde{\mathbf{x}}$, $\overline{\mathbf{x}}$,
$\widetilde{\mathbf{q}}$, $\overline{\mathbf{q}}$)
is used for both learning the projection-invariant encoding
and constructing the shared latent space between the two domains for
motion retargeting.

%
%
\subsection{Sampling Data from Kinematic Information} \label{subsec:sample_from_robot}
Here, we present a data sampling procedure 
which requires only the robot's kinematic information.
This has two major advantages.
1) it alleviates the cost of collecting 
a large number of expensive paired data obtained 
via solving inverse kinematics, and 2) it allows the learning-based motion retargeting algorithm to not be limited to the motion capture data used in
Section~\ref{subsec:sample_from_mocap}.

We first sample $\tilde{\mathbf{q}} \in \widetilde{\mathbb{Q}}$
where $\widetilde{\mathbb{Q}}$ is a relaxed robot configuration space. 
In particular, this relaxed configuration space is constructed by
relaxing the joint limits with a relaxation constant $\alpha$.
If the joint angle range of $i$-th joint is $[l_i, \, u_i]$, then we define the 
$\alpha$-relaxed joint range as 
$[ l_i - \frac{\alpha}{2}(u_i-l_i), \, u_i + \frac{\alpha}{2}(u_i-l_i)]$.
The relaxed configuration is converted to the corresponding
robot pose by solving forward kinematics where the skeleton features
(i.e., $7$ unit vectors) are extracted from the robot pose. 
We denote by
$\tilde{\mathbf{x}} = \texttt{FK}(\tilde{\mathbf{q}})$ 
the skeleton features given by the relaxed configuration.

The main purpose of sampling $\tilde{\mathbf{q}}$
from $\widetilde{\mathbb{Q}}$ is to 
collect $\tilde{\mathbf{x}}$ that is 
less restricted to the physical constraints of the robot model. 
In other words, if we simply use samples from the robot configuration
with exact joint ranges, the data is likely to
not cover the human skeleton space sufficiently.
Yet, if we directly use the collected relaxed pairs
$(\tilde{\mathbf{x}}, \, \tilde{\mathbf{q}})$ for training the 
learning-based motion retargeting, it may cause catastrophic results
in that $\tilde{\mathbf{q}}$ may not satisfy the kinematic constraints.

To resolve this issue, we use the same projection mapping from
Section~\ref{subsec:sample_from_mocap} to map the relaxed
configuration~$\tilde{\mathbf{q}}$
to the feasible robot configuration $\bar{\mathbf{q}}$
that is guaranteed to satisfy joint range limits and self-collision constraints
(i.e., $\bar{\mathbf{q}} = \texttt{Proj}_{\mathbb{Q}}( \tilde{\mathbf{q}} )$).
We also compute the corresponding skeleton feature from 
the feasible configurations
(i.e., $\bar{\mathbf{x}} = \texttt{FK}(\bar{\mathbf{q}})$).
To summarize, we sample the following tuples for training our 
learning-based motion retargeting method:
\begin{align*}
	\tilde{\mathbf{q}} & \sim P_{\widetilde{\mathbb{Q}}} \\
	\tilde{\mathbf{x}} & = \texttt{FK} (\tilde{\mathbf{q}}) \\
	\bar{\mathbf{q}} &= \texttt{Proj}_{\mathbb{Q}}( \tilde{\mathbf{q}} ) \\
	\bar{\mathbf{x}} & = \texttt{FK} (\bar{\mathbf{q}}) 
\end{align*}
With slight abuse of notation and denoting function composition by $\circ$,
we can represent 
$\bar{\mathbf{x}} = \texttt{Proj}_{\mathbb{X}}( \tilde{\mathbf{x}} )$,
where 
$\texttt{Proj}_{\mathbb{X}} = \texttt{FK} \circ 
\texttt{Proj}_{\mathbb{Q}} \circ \texttt{IK}$ even though we are not actually solving inverse kinematics in the 
sampling process.
Figure \ref{fig:mapping} illustrates the proposed projection-invariant mapping. 

We also leverage a mix-up style data augementaton~\cite{Zhang_18_mixup}.
More precisely, we generate auxiliary relaxed robot skeletons $\widetilde{\mathbf{x}}' = 
\beta \widetilde{\mathbf{x}} + (1-\beta)\overline{\mathbf{x}}$,
where $\beta$ is uniformly sampled from the $[0,1]$ interval.
This has the effect of smoothing the decision boundary
of the motion retargeting inputs.

%
%
\subsection{Discussion} \label{subsec:discussion}

We found that 200K pairs of skeleton features and
corresponding robot joint angles generated from 
the sampling method in Section \ref{subsec:sample_from_robot}
were not sufficient enough to properly perform motion retargeting.

This stems from the fact that 
the coverage of randomly sampled data becomes sparse at an exponential rate 
in the ambient dimension, which is typically referred to as the 
``curse of dimensionality''. 
One can try increasing the number of samples,
but the corresponding increase in query time would render it 
impractical for real-time retargeting when using 
exact nearest neighbor search. An avenue for future research is 
using sublinear-time algorithms for retrieving \emph{approximate} 
nearest neighbors.

Our method remedies the insufficient coverage of random data 
by using a small number of human motion capture data from
Section~\ref{subsec:sample_from_mocap} to 
efficiently cover the low-dimensional manifold of human skeletons.
However, we note that our method has the potential to be fully
\emph{self-supervised}, in the sense that no data obtained by solving inverse kinematics is required, if our random configuration sampling in
Section~\ref{subsec:sample_from_robot}
can be improved to efficiently cover the space of human
skeletons.

%
%
\section{Experiment}

%
%
\begin{figure*}[t!]
  \centering
  \includegraphics[width=0.8\textwidth]{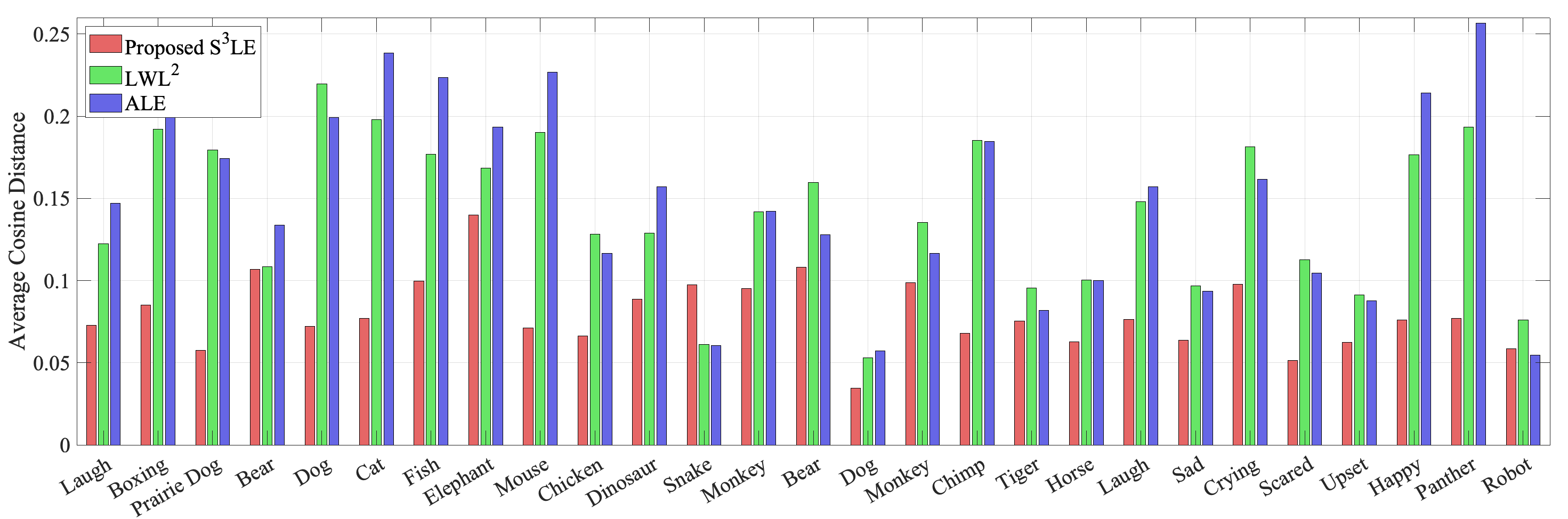}
  \vspace{-0.1pt}
  \caption{
    Average cosine distances between the retargeted and target 
    robot configuration of $27$ test motions. 
  }
  \label{fig:mr_res}
  \vspace{-0.1pt}
\end{figure*}

%
%
\begin{figure*}[t!]
  \centering
  \includegraphics[width=0.86\textwidth]{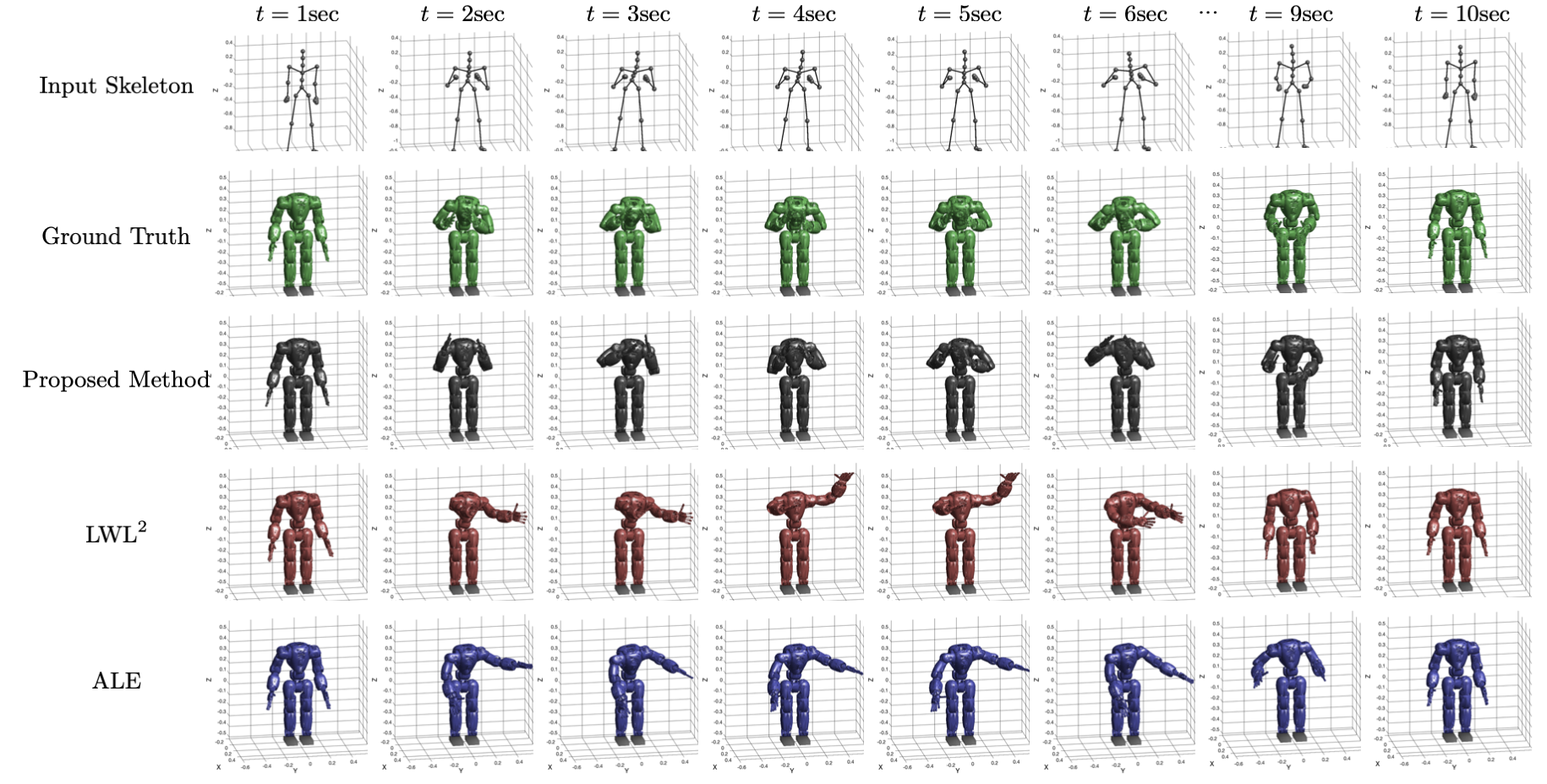}
  \vspace{-0.1pt}
  \caption{
    Motion retargeting results of the ground truth and
    compared methods on the \emph{Chimp} motion. 
  }
  \label{fig:mocap_exp}
  \vspace{-0.1pt}
\end{figure*}

\subsection{Synthetic Example}

We first conduct a synthetic experiment to conceptually highlight the benefit of using S$^3$LE for constructing the shared latent space.
The goal is to find a mapping from Relaxed~$X$ to Feasible~$Y$,
where multiple points in Relaxed~$X$ correspond to a single point in 
Feasible~$Y$. 
This resembles our motion retargeting setup since multiple
human skeletons can map into a single robot configuration 
due to the limited expressivity of humanoid robots.

We set Relaxed $X \subset \mathbb{R}^2$ and
Feasible $X \subset \mathbb{R}^2$ 
to be $[0,2]\times[0,2]$ and $[0,1]\times[0,1]$, respectively, and define
the projection as 
$\texttt{Proj}_{X}: (x_1,x_2) \mapsto (\min(x_1,1),\min(x_2,1))$. Note that there exists a one-to-one mapping from Feasible $X$
to Feasible $Y$ as shown in Figure \ref{fig:synthetic_transfer}.
We randomly sample $10,000$ points from Relaxed $X$ (which contains Feasible $X$ as a subset)
to optimize a projection-invariant mapping 
and $20$ pairs as the glue data for constructing the latent space shared by $X$ and $Y$.
The mappings learned by S$^3$LE, which is trained with the NT-Xent loss,
and a baseline method, trained without the NT-Xent loss, are illustrated in the
rightmost side of Figure \ref{fig:synthetic_transfer}. We can see that our method learns a better mapping
from Relaxed $X$ to Feasible $Y$.
This example shows that training with a contrastive loss 
improves the performance 
when the mapping is not one-to-one as the projection invariant 
region shown with red dotted circles are better mapped 
with the proposed method.

%
%
\begin{figure*}[t!]
  \centering
  \subfigure[]{\includegraphics[width=0.81\textwidth]
      {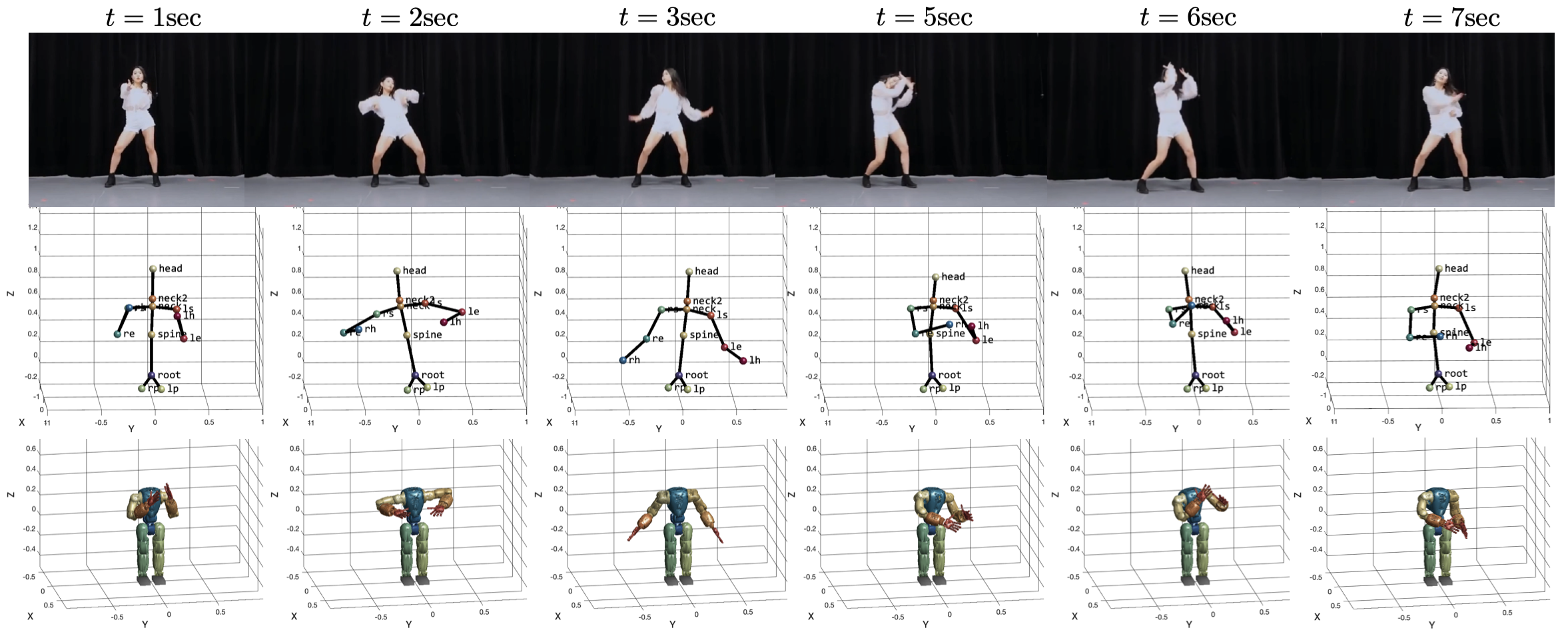}
      \label{fig:eft_coman_res_1}}
  \subfigure[]{\includegraphics[width=0.81\textwidth]
      {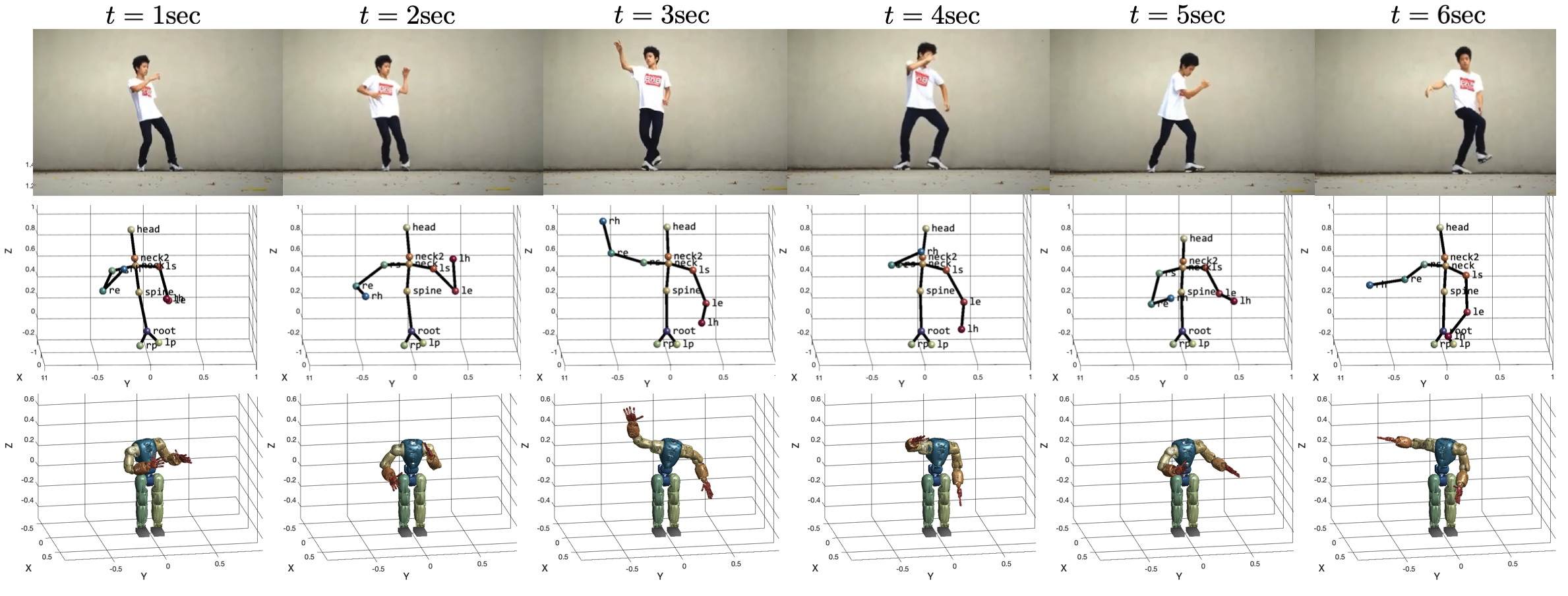}
      \label{fig:eft_coman_res_2}}     
  \vspace{-0.1pt}
  \caption{
  Snapshots of COMAN poses generated from two YouTube dancing videos.
  The second row and third row of each figure 
  indicate the upper-body of estimated human skeletons
  and retargeted robot configurations, respectively.
  The human skeleton is normalized in that 
  the line from the right pelvis to the left pelvis
  aligns with the $y$-axis. 
}
\label{fig:eft}
\vspace{-0.1pt}
\end{figure*}

\subsection{Experimental Setup} \label{subsec:setup}

For the target robot platform, 
Compliant Humanoid Platform (COMAN) 
is used in a simulational environment 
where we only consider the revolute joints of its upper body
and a subset of $10$ motions ($1,057$ pairs) from
CMU MoCap DB is used to collect 
a list of paired tuples
($\widetilde{\mathbf{x}}$, $\overline{\mathbf{x}}$,
$\widetilde{\mathbf{q}}$, $\overline{\mathbf{q}}$). 

We compare S$^3$LE
with two baseline methods, associate latent embedding (ALE)
\cite{Yin_17}
and locally weighted latent learning (LWL$^2$)
\cite{Choi_2020_RSS}. 
Both methods, ALE and LWL$^2$, leverage a shared latent space, but LWL$^2$ additionally guarantees feasibility of the retargeted pose using nonparametric regression. S$^3$LE also guarantees feasibility in the same way, i.e., by using nonparametric regression in the latent space.
Furthermore, the mixup-style augmentation in S$^3$LE yields more training data for constructing 
the shared latent space. 

We first select $10$ motions from the CMU MoCap database and 
run optimization-based motion retargeting to get $1,008$
pairs of skeletons and corresponding robot configurations. 
We also sample $100,000$ pairs from the proposed sampling process
in Section \ref{subsec:sample_from_robot},
$30,000$ from robot-domain specific data, and $100,000$ from MoCap-domain
specific data.
All neural networks (encoders, decoders, and discriminators)
have three hidden layers with $512$ units. We use
the $\text{tanh}$ activation function and learning rate of $10^{-3}$.

%
%
\subsection{Motion Retargeting with Motion Capture Database}

We conduct comparative experiments using
a subset of CMU MoCap DB~\cite{CMU_mocap} that was not observed while training as test set. 
In particular, we chose $27$ expressive motions, compute
the corresponding poses of COMAN using an optimization-based motion retargeting
\cite{Choi_19}, and use them as the ground truth.
To evaluate the performance of each retargeting method,
we compute the average cosine distance of seven unit vectors
(hip-to-neck, both left and right neck-to-shoulder,
shoulder-to-elbow, and elbow-to-hand)
between
the retargeted and ground truth skeletons. 
The average cosine distances of 
the methods are shown in Figure \ref{fig:mr_res}. In most cases, S$^3$LE outperforms
the baselines in terms of accuracy measured by the cosine distance. Note that the poses of both S$^3$LE and LWL$^2$ 
are guaranteed to be collision-free while the average collision rate of ALE is
$24.9\%$.
Figure \ref{fig:mocap_exp} shows snapshots of input skeleton poses
and retargeted motions. S$^3$LE accurately retargets the motion when the input skeleton is lifting both hands up, whereas the baseline methods fail to do so.
We hypothesize that this is because the training data from optimization-based motion retargeting failed to cover such 
poses while the data sampled using the COMAN's kinematic information (Section \ref{subsec:sample_from_robot}) contained such poses.

%
%
\subsection{Motion Retargeting with RGB Video}

We further show that the proposed method can be used to generate natural and stylistic robotic motions
from RGB videos. We use FrankMoCap 
\cite{Rong_20_frankmocap}, the state-of-the-art $3$-D pose estimation algorithm, to collect sequences of human $3$-D skeleton poses from YouTube videos. 
Since there is no ground truth, we simply show
snapshots of reference RGB images and corresponding
COMAN poses in Figure \ref{fig:eft} for qualitative evaluation. 
More videos can be found in the supplementary materials.

%
%
\section{Conclusion}
In this paper, we presented S$^3$LE, a self-supervised motion retargeting method which consists of two parts: 1) collecting tuples of relaxed and retargetable skeleton poses, and corresponding robot configurations to reduce the necessity of expensive data obtained via solving inverse kinematics,
and 2) learning projection-invariant encoders to handle the different expressivities of humans and humanoid robots while constructing the shared latent space.
Furthermore, our method guarantees the safety of the robot motion by utilizing a nonparametric method in the shared latent space.
We demonstrate that S$^3$LE can efficiently generate 
expressive and safe robotic motions from both motion capture data
and YouTube videos.

One downside of using the nonparametric method
is that its computational complexity increases linearly 
in the number of data. This becomes problematic 
when we want to retarget motions in real-time with a large number of pre-stored poses. 
We plan to investigate sublinear-time approximate nearest neighbor
algorithms~\cite{Indyk_98_ann,Charikar_02_hyperplanelsh,Andoni_18_annsurvey}
(e.g., locality sensitive hashing) to speed-up the inference.

\balance

%
%
\bibliographystyle{IEEEtran}
\bibliography{references}

\end{document}